\documentclass[runningheads]{llncs}
\usepackage{graphicx}
\begin{document}
\title{RoboKit-MV: an Educational Initiative}
%


%
\author{
Azer Babaev \and
Ilya Osokin \and
Ilya Ryakin \and
Egor Davydenko \and
Vladimir Litvinenko \and
Ivan Khokhlov \and
Aleksandr Matsun \and
Vitaly Suvorov}

\authorrunning{A. Babaev et al.}
%

\institute{Team Starkit, Moscow Institute of Physics and Technology, Russia
\email{robocup.mipt@gmail.com}\\
}
\maketitle              
\begin{abstract}
In this paper, we present a robot model and code base for affordable education in the field of humanoid robotics. We give an overview of the software and hardware of a robot that won several competitions with the team RoboKit in $2019-2021$, provide analysis of the contemporary market of education in robotics, and highlight the reasoning beyond certain design solutions.

\end{abstract}

\section{Introduction}

Autonomous artificial humanoid beings have fascinated people since ancient times.
From the sculpture, animated in the myth of Pygmalion, to golem, they have remained to be a common trope for centuries.
And in the \textsc{XX} century they were finally embedded in the form of moving machines, yet with metal and code instead of the clay and parchment.

\subsection{Contemporary robotics}

In recent years certain trends in the market and the industry of robotics are becoming more and more visible.

First, with the development of the high-torque electric actuators, such as \cite{mit_servos}, \cite{odri} the robots are becoming more dynamical.
These actuators allow them not only to move at high speeds, but also to rapidly create forces and torques to perform dynamic actions, such as running, jumping, etc.

Second, robotics is moving towards legged solutions.
While wheeled robots are generally simpler to produce and control, legged ones have wider applicability in terms of terrain requirements.
Quadrupedal platforms nowadays find real industrial applications \cite{quadrupedal_industrial}.

Speaking of the bipedal solutions, it is worth noting that they are typically much more complex than the wheeled ones in terms of both mechanics and control algorithms.
A minimal wheeled robot requires $2$ servomotors, a typical biped (legs only) consists of $12$.
The body with two fully controlled feet is an omnidirectional platform, capable of following a trajectory in three dimensions.

Third, the overall growth of the on-board computational capabilities gives even the relatively small, low-power robots \cite{starkit_champion} an opportunity to perform computationally expensive operations in real time.
They include classical CV-based or NN-based point cloud and depth analysis, Particle Filters, SLAM, NN detectors inference, and other computationally intensive operations.

\subsection{The state of educational robotics for school students in Russia}

In $2021$ so called ''robotics for children'' could be typically classified into one of the following.

First, Lego-like toys, Scratch programming, visual algorithms.
It is beyond necessity to unwrap how far is it from any industrial applications or real-world robotics.
These approaches are suitable for introducing robotics to children, make children familiar with the core concepts of building a robotic system, but not more than that.

The second group consists of such robots as Bioloid, Abilix and their analogues.
In certain modifications, they are humanoid or nearly humanoid.
But weak servomotors with plastic reduction gears, outdated cameras with resolution of nearly $200 \times 200$ and insufficient computational powers limit the range of possible actions for the robot.
Harsh backlashes in those servos make it impossible to perform dynamic movements, and complex vision cannot run on microcontrollers.

The next type of the robots that are used in education (rarely because of their price) includes Darwin, Nao and the same kind of expensive multi-functional humanoids.
Darwin requires student-level expertise, it is often used in robotic laboratories and competitions.
Nao robot is easier to start with, but it also has certain shortcomings, such as fragile fingers, relatively weak hand servos, no room for customization.
High utilization of its capabilities (as it is done in RoboCup SPL) requires strong knowledge of ROS, networks.

Overall, the mentioned means of education do not meet the market demands in terms of robot capabilities, simplicity of programming, and price.

\section{Solution overview}

We propose using a robot capable of playing football as an educational platform.
Robotic football is a complex, dynamic, well-examined multiagent game.
Starting from $1996$, humanoid and non-humanoid robots are competing in several leagues in bringing the ball to the opponent goal, just like in the human football.
These factors make it possible for such a robot to take part in another competitions, such as sprint, marathon, obstacle avoidance, and other.

\subsection{Software}

The controlling software of the robot is implemented in $MicroPython$, the same language is supposed to be used by the school student to program it.
The smart camera $OpenMV$ has built-in libraries for image processing, making it possible to implement vision in the absence of $OpenCV$ and $numpy$.
The code is running in a single thread, which is enough for educational purposes.
First, it makes it possible to debug the code linearly, and secondly, it naturally makes it necessary to optimize the computations.

\medskip

The code base consists of the following modules.

\medskip

\textbf{Vision} module contains the application of the classical Image Processing operations, such as color-based mask obtainment, shape and connected components analysis, as well as the analysis of the geometrical relations between certain objects in the frame. Treating the image as an array, it gives the pixel coordinates of the objects as an output.

\medskip

\textbf{Model} is responsible for mapping the pixel coordinates of the objects of interest to the real-world coordinates in meters. It uses the calibration matrix of the camera, and via utilizing a pinhole camera model, such a transform is performed. At this point the objects remain in the robot's coordinate system.

\medskip

\textbf{Localization} takes the output of the previous processing stage and performs the inverse operation of calculating the position of the robot in the field that will fit the observations best. The approach that is used relies on the Particle Filter, because it provides sufficient robustness under the noisy measurements. Moreover, the Localization module performs the filtering of the ball observations in order to extract its position from the data as precise as possible.

\medskip

\textbf{Strategy} module is responsible for choosing the next operation to perform, given the current field state. Since the end goal of the robot is to place the ball in the opponent's goal, it generates a trajectory, that starts in the current ball coordinates and ends in the middle of the goal. The trajectory is generated from the scratch on each step to compensate for the changing environment, noise in input and the actions of the opponent robots.

\medskip

\textbf{Motion} is the last module in the pipeline, performing the planned actions. At the early stages of development it heavily relied on the pre-recorded motions, inherited from the original robot, now it is mainly implemented on the basis of the inverse kinematics module. The walking pattern that is used is semi-dynamic in a sense that the robot is exploiting swing motion from side to side while walking, not being in a stable equilibrium most of the time. The inverse kinematics is calculated on a single-core $Cortex$ processor in $1$ $ms$.

\subsection{Hardware}


We have used a remotely controlled humanoid robot Kondo \cite{kondo} as a starting point, and modified it for out needs. We have integrated a smart camera $OpenMV$ in the robot, replacing its head. We have modified its pelvis construction to increase the range of possible motions.

\begin{figure}
    \centering
    \includegraphics[width=0.7\textwidth]{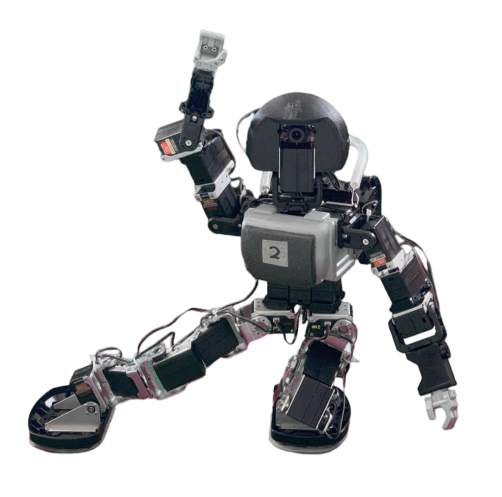}
    \label{fig:field}
    \caption{$23$-DoF humanoid robot RoboKit-MV with the computer installed in the head. Note the range of the degrees of freedom in the feet.}
\end{figure}

The robot contains $23$ servomotors: $17$ $KRS-2552RHV$ servos and $6$ servos
$KRS-2572HV$ in the legs. $KRS-2552RHV$ servos have maximal torque of $1,37$ ${N}\cdot{m}$
and Max Speed $7.48$ $rad/s$. Also, $KRS-2552RHV$ servos has Max Torque $2,45$ ${N}\cdot{m}$ and Max Speed $8.05$ $rad/s$. The battery for $1300 mAh$ is used as a power source, which allows the robot to play for about $10$ minutes autonomously.

\section{Educational use and accomplishments}

The educational use of this robot does not imply any necessary low-level programming.
The first problem that is supposed to be solved by the student is to set a strategy in the graphic form: for each section of the field, one should assign a desired direction of the ball movement in order to score a goal.
After the student has mastered this problem, Python programming comes into place.
Vision have shown itself as a good starting point to program a robot, since it does not necessarily require the hardware all the time, and on the basic level it is a set of pretty straightforward processing methods.

The proposed platform proved itself in $RoboCup$ and $FIRA$ competitions. Thir robot is the champion of $RoboCup$ $Asia$ $Pacific$ games in $2019$ and $2020$. Moreover, $RoboKit-MV$ set the world record among juniors in triple jump and won in sprint and marathon.

\section{Future work}

We are planning to make the following changes to the robot:

\begin{itemize}
    \item introduce stereo vision, which was an important feature in another our robot \cite{starkit_champion}
    \item increase the $DoF$ number for upper limbs
    \item introduce custom $STM-32$ controller to further improve the quality and the bandwidth of the low-level control
    \item switch to the $x86$ computer
    \item increase the battery capacity
\end{itemize}



%
%
%
%

\end{document}